\documentclass[runningheads]{llncs}
\usepackage{graphicx}
\usepackage{amsmath,amssymb} %
\usepackage{color}
\usepackage[width=122mm,left=12mm,paperwidth=146mm,height=193mm,top=12mm,paperheight=217mm]{geometry}

\usepackage[pagebackref=true,breaklinks=true,letterpaper=true,colorlinks,bookmarks=false]{hyperref}
\usepackage{subfig}
\usepackage{multirow}

\begin{document}
\pagestyle{headings}
\mainmatter

\definecolor{olive}{RGB}{50,150,50}

\newif\ifdraft

\ifdraft
 \newcommand{\PF}[1]{{\color{red}{\bf pf: #1}}}
 \newcommand{\pf}[1]{{\color{red} #1}}
 \newcommand{\HR}[1]{{\color{blue}{\bf hr: #1}}}
 \newcommand{\hr}[1]{{\color{blue} #1}}
 \newcommand{\VC}[1]{{\color{green}{\bf vc: #1}}}
  \newcommand{\vc}[1]{{\color{green} #1}}
 \newcommand{\ms}[1]{{\color{olive}{#1}}}
 \newcommand{\MS}[1]{{\color{olive}{\bf ms: #1}}}
 \newcommand{\JS}[1]{{\color{cyan}{\bf js: #1}}}
 \newcommand{\NEW}[1]{{\color{red}{#1}}}

\else
 \newcommand{\PF}[1]{{\color{red}{}}}	
 \newcommand{\pf}[1]{ #1 }
 \newcommand{\HR}[1]{{\color{blue}{}}}
 \newcommand{\hr}[1]{#1}%
 \newcommand{\VC}[1]{{\color{green}{}}}
 \newcommand{\ms}[1]{ #1 }
 \newcommand{\MS}[1]{{\color{olive}{}}}
\fi

\newcommand{\vp}{\mathbf{p}}
\newcommand{\mI}{\mathbf{I}}
\newcommand{\mB}{\mathbf{B}}
\newcommand{\mR}{\mathbf{R}}

\newcommand{\cA}{\mathcal A}
\newcommand{\cC}{\mathcal C}
\newcommand{\cD}{\mathcal D}
\newcommand{\cE}{\mathcal E}
\newcommand{\cF}{\mathcal F}
\newcommand{\cL}{\mathcal L}
\newcommand{\cU}{\mathcal U}

\newcommand{\TODO}[1]{\textcolor{red}{#1}}
\newcommand{\R}{\mathbb{R}}
\newcommand{\Latent}{\mathbf{L}}
\newcommand{\LatentG}{\Latent^{\text{3D}}} %
\newcommand{\LatentA}{\Latent^\text{app}} %
\newcommand{\LatentBG}{\mB} %
\newcommand{\fr}{t}
\newcommand{\pcentroid}{\hat{p}}

\newcommand{\comment}[1]{}

\newcommand{\argmin}{\operatornamewithlimits{argmin}}

\newcommand{\unet}[0]{{\bf OursUnet}}
\newcommand{\rnet}[0]{{\bf OursResnet}}
\newcommand{\rhod}[0]{{\bf Rhodin}}
\newcommand{\resn}[0]{{\bf Resnet}}

\title{Unsupervised Geometry-Aware Representation for 3D Human Pose Estimation}

\titlerunning{Unsupervised Geometry-Aware Representation}

\authorrunning{Helge  Rhodin \and Mathieu Salzmann \and Pascal  Fua}

\author{Helge  Rhodin \and Mathieu Salzmann \and Pascal  Fua}
\institute{CVLab, EPFL, Lausanne, Switzerland}

\maketitle

\begin{abstract} %

Modern 3D human pose estimation techniques rely on deep networks, which require large amounts of training data. While weakly-supervised methods require less supervision, by utilizing 2D poses or multi-view imagery without annotations, they still need a sufficiently large set of samples with 3D annotations for learning to succeed. 

In this paper, we propose to overcome this problem by learning a geome\-try-aware body representation from multi-view images without annotations. To this end, we use an encoder-decoder that predicts an image from one viewpoint given an image from another viewpoint. Because this representation encodes 3D geometry, using it in a semi-supervised setting makes it easier to learn a mapping from it to 3D human pose. As  evidenced by our experiments, our approach significantly outperforms fully-supervised methods given the same amount of labeled data, and improves over other semi-supervised methods while using as little as 1\% of the labeled data.

\end{abstract}

\section{Introduction}

Most current monocular solutions to 3D human pose estimation rely on methods based on convolutional neural networks (CNNs). With networks becoming ever more sophisticated, the main bottleneck now is the availability of sufficiently large training datasets, which typically require a large annotation effort. While such an effort might be practical for a handful of subjects and specific motions such as walking or running, covering the whole range of human body shapes, appearances, and poses is infeasible.

Weakly-supervised methods that reduce the amount of annotation required to achieve a desired level of performance are therefore valuable.  For example, methods based on articulated 3D skeletons can be trained not only with actual 3D annotations but also using 2D annotations~\cite{Mehta17a,Zhou17a} and multi-view footage \cite{Pavlakos17,Tung17self}. Some methods dispense with 2D annotations altogether and instead exploit multi-view geometry in sequences acquired by synchronized cameras~\cite{Rhodin18a,Zhou17unsupervised}. However, these methods still require a good enough 3D training set to initialize the learning process, which sets limits on the absolute gain that can be achieved from using unlabeled examples.

\begin{figure}[t!]
	\centering
	\includegraphics[width=0.9\linewidth]{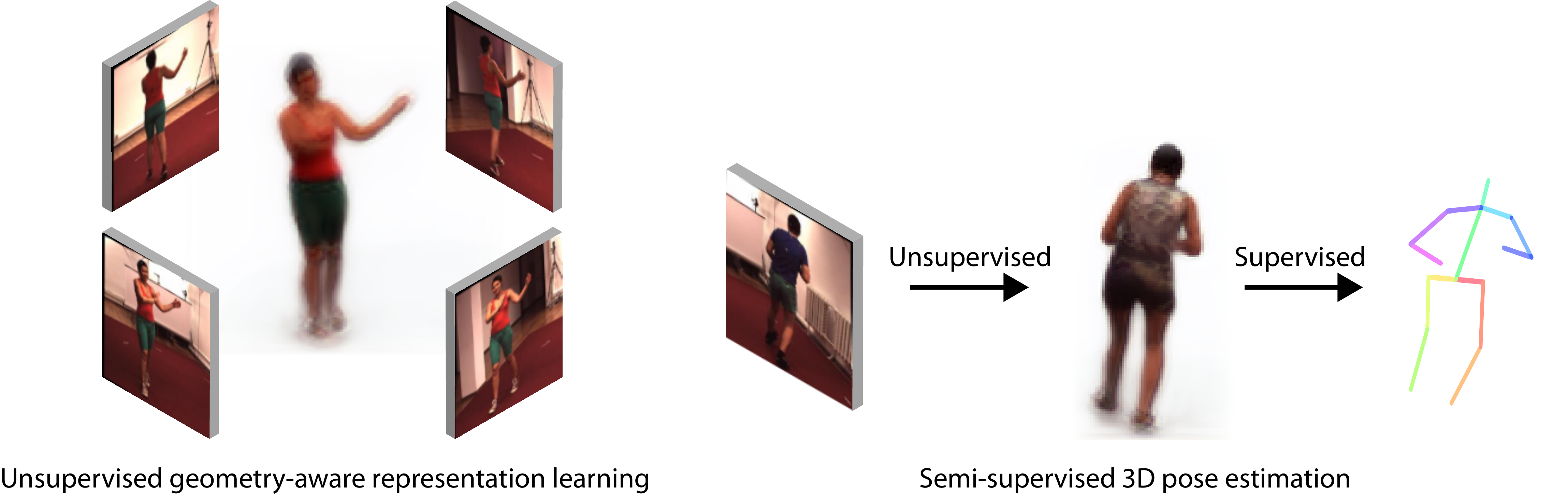}\\
	(a) \\[2mm]
	\begin{tabular}{cc}
	\includegraphics[width=0.48\linewidth]{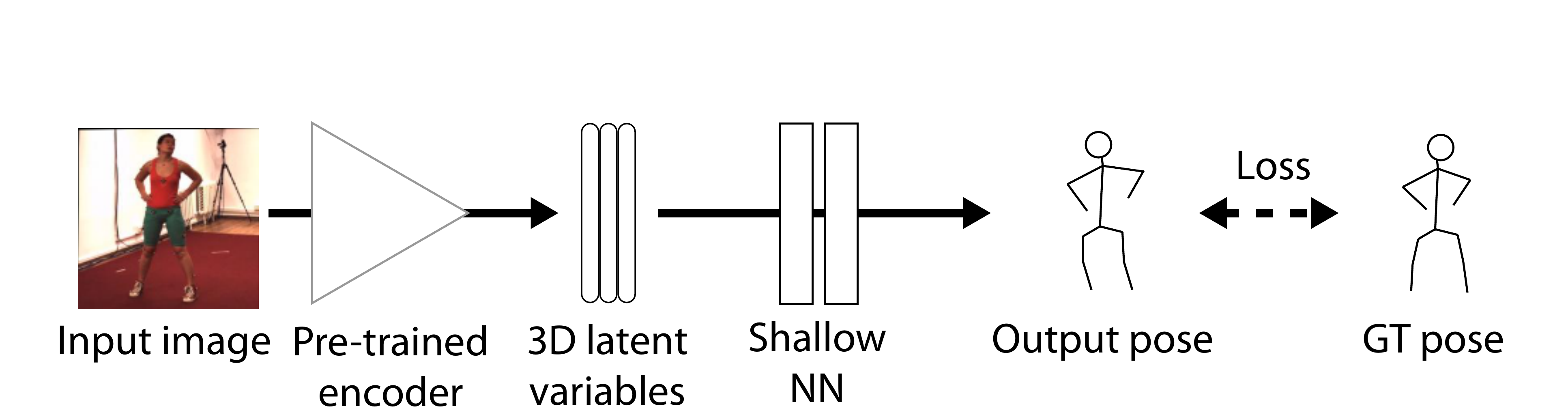}&
	\includegraphics[width=0.48\linewidth]{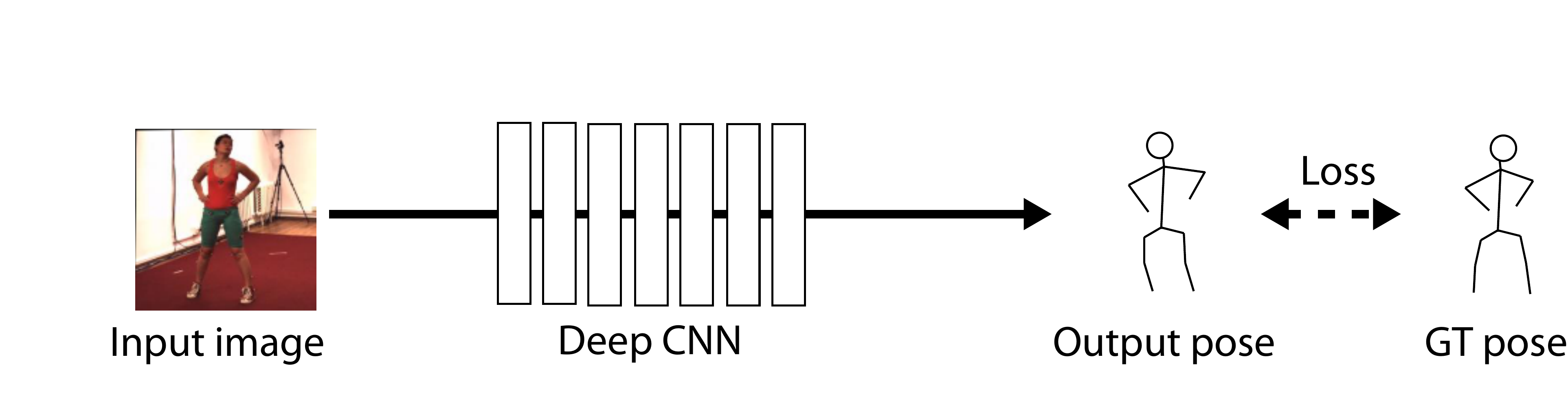}\\
	(b)&(c)
	\end{tabular}
	\vspace{-3mm}
	\caption{{\bf Approach.} {\bf (a)} During training, we first learn a geometry-aware representation using unlabeled multi-view images. We then use a small amount of supervision to learn a mapping from the our representation to actual 3D poses, which only requires a shallow network and therefore a limited amount of supervision.  {\bf (b)} At run-time, we compute the latent representation of the test image and feed it to the shallow network to compute the pose.  {\bf (c)} By contrast, most state-of-the-art approaches train a network to regress directly from the images to the 3D poses, which requires a much deeper network and therefore more training data.}
	\label{fig:teaser}
\end{figure}

In this paper, we propose to use images of the same person taken from multiple views to learn a latent representation that, as shown on the left side of Fig.~\ref{fig:teaser}(a), captures the 3D geometry of the human body. Learning this representation does not require any 2D or 3D pose annotation. Instead, we train an encoder-decoder to predict an image seen from one viewpoint from an image captured from a different one. As sketched on the right side of Fig.~\ref{fig:teaser}(a), we can then learn to predict a 3D pose from this latent representation in a supervised manner. The crux of our approach, however, is that because our latent representation already captures 3D geometry, the mapping to 3D pose is much simpler and can be learned using much fewer examples than existing methods that rely on multiview supervision~\cite{Rhodin18a,Zhou17unsupervised}, and more generally most state-of-the-art methods that attempt to regress directly from the image to the 3D pose.

As can be seen in Fig.~\ref{fig:teaser}, our latent representation resembles a volumetric 3D shape. While such shapes can be obtained from silhouettes~\cite{Yan16,Tulsiani17}, body outlines are typically difficult to extract from natural images. By contrast, learning our representation does not require any silhouette information. Furthermore, at test time, it can be obtained from a monocular view of the person. Finally, it can also be used for novel view synthesis (NVS) and outperforms existing encoder-decoder algorithms~\cite{Tatarchenko15,Tatarchenko16,Park17} qualitatively on natural images. 

Our contribution is therefore a latent variable body model that can be learned without 2D or 3D annotations, encodes both 3D pose and appearance, and can be integrated into semi-supervised approaches to reduce the required amount of supervised training data.  We demonstrate this on the well-known Human3.6Million~\cite{Ionescu14a} dataset and show that our method drastically outperforms fully supervised methods in 3D pose reconstruction accuracy when only few labeled examples are available.

\section{Related work}
In the following, we first review the literature on semi-supervised approaches to monocular 3D human pose estimation, which is most closely related to our goal. We then discuss approaches that, like us, make use of geometric representations, both in and out of the context of human pose estimation, and finally briefly review the novel view synthesis literature that has inspired us.

\vspace{-3mm}
\paragraph{\bf  Semi-supervised human pose estimation.}

While most current human pose estimation methods~\cite{Pavlakos17,Zhou17a,Pavlakos16,Tome17,Popa17,Martinez17,Mehta17b,Rogez17,Tekin17a} 
are fully supervised, relying on large training sets annotated with ground-truth 3D positions coming from multi-view motion capture systems~\cite{Mehta17a,Ionescu14b}, several methods have recently been proposed to limit the requirement for labeled data.
In this context, foreground and background augmentation~\cite{Rhodin16,Rogez16} and the use of synthetic datasets \cite{Chen16,Varol17} focus on increasing the training set size. Unfortunately, these methods do not generalize well to new motions, apparels, and environments that are different from the simulated data. Since larger and less constrained datasets for 2D pose estimation exist, they have been used for transfer learning \cite{Tung17self,Mehta17b} and to provide re-projection constraints~\cite{Zhou17a}. Furthermore, given multiple views of the same person,  3D pose can be triangulated from 2D detections~\cite{Pavlakos17,Joo15} and a 2D pose network can be trained to be view-consistent after bootstrapping from annotations. Nevertheless, these methods still require 2D annotation in images capturing the
target motion and appearance. By contrast, the methods of~\cite{Rhodin18a,Zhou17unsupervised} exploit multi-view geometry in sequences acquired by synchronized cameras, thus removing the need for 2D annotations. However, in practice, they still require a large enough 3D training set to initialize and constrain the learning process. We will show that our geometry-aware latent representation learned from multi-view imagery but without annotations allows us to train a 3D pose estimation network using much less labeled data.

\vspace{-3mm}
\paragraph{\bf Geometry-aware representations.}

Multi-view imagery has long been used to derive volumetric representations of 3D human pose from silhouettes, for example by carving out the empty space. This  approach can be used in conjunction with learning-based methods~\cite{Tulsiani18}, by defining constraints based on perspective view rays~\cite{Tulsiani17,Kar17}, orthographic projections~\cite{Yan16}, or learned projections~\cite{Rezende16}. 
It can even be extended to the single-view training-scenario if the distribution of the observed shape can be inferred prior to reconstruction~\cite{Zhu17,Gadelha16}. The main drawback of these methods, however, is that accurate silhouettes are difficult to automatically extract in natural scenes, which limits their applicability. 

Another approach to encoding geometry relies on a renderer that generates images from a 3D representation \cite{Grant16,Shu17a,Kim17,Zhao17} and can function as a decoder in an autoencoder setup~\cite{Bas17,Tewari17}. For simple renderers, the rendering function can even be learned~\cite{Dosovitskiy15,Dosovitskiy17} and act as an encoder. When put together, such learned encoders and decoders have been used for unsupervised learning, both with GANs \cite{Chen16info,Tran17,Tung17} and without them~\cite{Kulkarni15}. In~\cite{Thewlis17,Thewlis17equivariant}, a CNN was trained to map to and from spherical mesh representations without supervision. While these methods also effectively learn a geometry-aware representation based on images, they have only been applied to well-constrained problems, such as face modeling. As such, it is unclear how they would generalize to the much larger degree of variability of 3D human poses. 

\comment{
\paragraph{Depth and other maps.}

Shapes have also been inferred by regressing silhouette, depth, and normal maps from opposing views \cite{Soltani17,Wu17marrnet,Lin17learning,Wiles17}.
CNNs can be trained to predict a depth map from a monocular image by utilizing photo- and left-right-consistency in stereo recordings \cite{Garg16,Godard17}.
Even monocular unsupervised training is possible by predicting the camera pose between two views of the same scene alongside depth and using re-projection photo-consistency \cite{Zhou17unsupervised}.
While it is less expensive to encode label assignments in 2D maps \cite{VNECT} instead of discrete 3D volumes, they are view-dependent and can only map a projection of the 3D object, furthermore, \cite{Zhou17unsupervised} requires a static scene..

\paragraph{Surface meshes.}

Meshes form a compact and complete surface representation and XXXXXXX, however, the mesh topology must be predefined prior to training with a template mesh \cite{Kurenkov17}.
\HR{TODO: check again} %
\cite{Rezende16}.
\cite{Thewlis17} Unsupervised learning of object landmarks by factorized spatial embeddings
\cite{Thewlis17equivariant} Unsupervised learning of object frames by dense equivariant image labelling
}

\comment{
\paragraph{\bf Rendering models.}

If training is performed on rendered images of a parametric model, the association of pose, shape and illumination parameters is known and their relation can be learned \cite{Grant16,Shu17a,Kim17,Zhao17}.
In effect, the renderer establishes the relation between model and image. On real images, the renderer can function as a decoder in an autoencoder setup~\cite{Bas17,Tewari17}. For simple renderers, the rendering functions can be learned \cite{Dosovitskiy15,Dosovitskiy17} and act as an encoder. In combination, such learned encoders and decoders have been used for unsupervised learning, with GANs \cite{Chen16info,Tran17,Tung17} and without \cite{Kulkarni15}. In~\cite{Thewlis17,Thewlis17equivariant}, a CNN is trained to map to and from spherical mesh representations without supervision , which is well suited for faces that unfold well onto a sphere, but has not been applied to non-concave cases such as articulated human pose.
}

\vspace{-3mm}
\paragraph{\bf Novel view synthesis.}

Our approach borrows ideas from the novel view synthesis literature, which is devoted to the task of creating realistic images from previously unseen viewpoints. Most recent techniques rely on  encoder-decoder architectures, where the latent code is augmented with view change information, such as yaw angle, and the decoder learns to reconstruct the encoded image from a new perspective~\cite{Tatarchenko15,Tatarchenko16}. Large view changes are difficult. They have been achieved by relying on a recurrent network that performs incremental rotation steps~\cite{Yang15weakly}. 
Optical flow information~\cite{Park17,Zhou16view} and depth maps~\cite{Flynn16} have been used to further improve the results. While the above-mentioned techniques were demonstrated on simple objects, methods dedicated to generating images of humans have been proposed. However, most of these methods use additional information as input, such as part-segmentations~\cite{Lassner17generative} and 2D poses~\cite{Ma17}. Here, we build on the approaches of~\cite{Cohen14,Worrall17} that have been designed to handle large viewpoint changes. We describe these methods and our extensions in more detail in Section~\ref{sec:latent}.

\comment{
\paragraph{Transformation network architectures.}

Special neural network architectures have been developed that explicitly model rotation \cite{Worrall17harmonic} and affine transformations \cite{Hinton11,Jaderberg15,Sabour17}.
These developments are orthogonal and have the potential to improve our simple but general encoder and decoder. %
}

\comment{
*Disentangling representations*
Learning Disentangled Representations with Semi-Supervised Deep Generative Models (distantly related?)

*Generalized forms of rendering/projection*
XXXXXXXXXXXXXXX (Task specific 'renderers' e.g. rotation and projection for 3D pose, to compare it with 2D pose. Not fully unsupervised, but 2D pose is provided.)
Single Image 3D Interpreter Network (weakly supervised on 2D pose, using 3D models as constraints)

*Representation learning*
Representation Learning: A Review and New Perspectives ( A good overview, quite general about autoencoders..)
Deep Learning of Representations: Looking Forward (similar overview)

*Volumetric representations*
??? Projective generative adversarial networks (PrGANs, trains a deep generative model of 3D shapes whose projections match the distributions of the input 2D views. without using any 3D, viewpoint information, or annotation during the learning phase)

*Distantly related*
Image-to-Image Translation from Unpaired Supervision (map image of horses to image of zebras, without having explicit/spatial correspondence in the image)
GENERATING INTERPRETABLE IMAGES WITH CONTROLLABLE STRUCTURE (word description as well as image sketch/2D joint locations to synthesize an image)
DNA-GAN: LEARNING DISENTANGLED REPRESENTATIONS FROM MULTI-ATTRIBUTE IMAGES (represenations which can be added and subtracted to yield new faces)
Attribute2Image: Conditional Image Generation
from Visual Attributes (Generate images given description)

*Predicting the future from video and or stereo*
Unsupervised Learning of Disentangled Representations from Video
Learning to Generate Long-term Future via Hierarchical Prediction (forecoast heatmap, then forecast image from heatmap)
}

\section{Unsupervised Geometry-Aware Latent Representation}
\label{sec:latent}

Our goal is to design a latent representation $\Latent$ that encodes 3D pose, along with shape and appearance information, and can be learned without any 2D or 3D pose annotations. To achieve this, we propose to make use of  sequences of images acquired from multiple synchronized and calibrated cameras. To be useful, such footage requires care during the setup and acquisition process. However, the amount of effort involved is negligible compared to what is needed to annotate tens of thousands of 2D or 3D poses.

For $\Latent$ to be practical, it must be easy to decode into its individual components. To this end, we learn from the images separate representations for the body's 3D pose and geometry, its appearance, and that of the background. We will refer to them as $\LatentG$, $\LatentA$, and $\LatentBG$, respectively. 

Let us assume that we are given a set, $\cU = \{ (\mI_t^i,\mI_t^j )\}_{t=1}^{N_u}$,  of $N_u$ image pairs without annotations, where the $i$ and $j$ superscripts refer to the cameras used to capture the images, and the subscript $t$ to the acquisition time. Let $\mR^{i \to j}$ be the rotation matrix from the coordinate system of  camera $i$ to that of camera $j$. We now turn to the learning of the individual components of $\Latent$.

\begin{figure}[t]
	\centering
	\includegraphics[width=\linewidth]{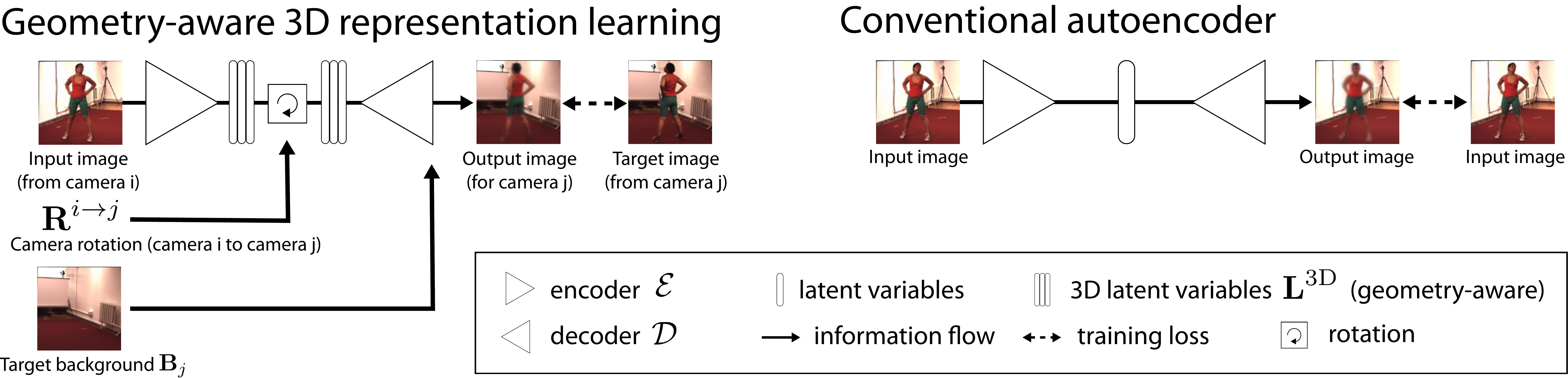}
\caption{{\bf Representation learning.} We learn a representation that encodes geometry and thereby 3D pose information in an unsupervised manner. Our method (Left) extends a conventional auto encoder (Right) with a 3D latent space, rotation operation, and background fusion module. 
	The 3D rotation enforces explicit encoding of 3D information. The background fusion enables application to natural images. }
	\label{fig:representationLearning}
\end{figure}

\paragraph{\bf Learning to encode multi-view geometry.} 

For individual images, autoencoders such as the one shown on the right side of Fig.~\ref{fig:representationLearning} have become standard tools to learn latent representations in unsupervised settings. Let such an autoencoder comprise an encoder $\cE_{\theta_e}$ and a decoder  $\cD_{\theta_d}$, where $\theta_e$ and $\theta_d$ are the weights controlling their behaviors. For image representation purposes, an autoencoder can be used to encode an image $\mI$ into a  latent representation $\Latent=\cE_{\theta_e}(\mI)$, which can then be decoded into a reconstructed image $\hat{\mI}=\cD_{\theta_d}(\Latent)$. $\theta_e$ and $\theta_d$ are learned by minimizing $\| \mI-\hat{\mI} \|^2$ on average over a training set $\cU$.

To leverage multi-view geometry, we take our inspiration from Novel View Synthesis methods~\cite{Tatarchenko15,Tatarchenko16,Cohen14,Worrall17} that rely on training encoder-decoders on multiple views of the same object, such as a car or a chair.  Let $(\mI_t^i, \mI_t^j) \in \cU$ be two images taken from different viewpoints but at the same time $t$. Since we are given the rotation matrix  $\mR^{i \to j}$ connecting the two viewpoints, we could feed this information as an additional input to the encoder and decoder and train them to encode $\mI_t^i$ and resynthesize  $\mI_t^j$, as in~\cite{Tatarchenko15,Tatarchenko16}. 
Then, novel views of the object 
could be rendered by varying the rotation parameter $\mR^{i \to j}$.
However, this does not force the latent representation to encode 3D information explicitly.
To this end, we model the latent representation
$\LatentG \in \R^{3\times N}$ as a set of $N$ points in 3D space
by designing the encoder $\cE_{\theta_e}$ and decoder $\cD_{\theta_e}$ so that they have a three channel output and input, respectively, as shown on the left side of Fig.~\ref{fig:representationLearning}. 
This enables us to model the view-change as a proper 3D rotation by matrix multiplication of the encoder output by the rotation matrix before using it as input to the decoder. Formally, the output of the resulting autoencoder $\cA_{\theta_e,\theta_d}$ can be written as
\begin{equation}
\cA_{\theta_e,\theta_d}(\mI_t^i, \mR^{i \to j}) =  \cD_{\theta_d}(\mR^{i \to j} \LatentG_{i,t} ) \text{, with } \LatentG_{i,t} = \cE_{\theta_e}(\mI_t^i) \; ,
\label{eq:latentG}
\end{equation}
and the weights $\theta_d$ and $\theta_e$ are optimized to minimize $\| \cA_{\theta_e,\theta_d}( \mI_t^i, \mR^{i \to j})  - \mI_t^j \|$ in the least-square sense over the training set $\cU$. 
In this setup, which was also used in~\cite{Cohen14,Worrall17},  the decoder $\cD$ does not need to learn how to rotate the input to a new view but only how to decode the 3D latent vector $\LatentG$. This means that the encoder is forced to map to a proper 3D latent space, that is, one that can still be decoded by $\cD$ after an arbitrary rotation. However, while $\LatentG$ now encodes multi-view geometry, it also encodes the background and the person's appearance. Our goal now is to isolate them from $\LatentG$ and to create two new vectors $\LatentBG$ and $\LatentA$ that encode the latter two so that  $\LatentG$ only represents geometry and 3D pose.

\paragraph{\bf Factoring out the background.}

Let us assume that we can construct background images $\mB_j$, for example by taking the median of all the images taken from a given viewpoint $j$. To factor them out, we introduce in the decoder a direct connection to the target background $\mB_j$, as shown in Fig.~\ref{fig:representationLearning}. More specifically, we concatenate the background image with the output of the decoder and use an additional $1\times 1$ convolutional layer to synthesize the decoded image.  This frees the rest of the network from having to learn about the background and ensures that the $\LatentG$ vector we learn does not contain information about it anymore.

\begin{figure}[t]
	\centering
	\includegraphics[width=\linewidth]{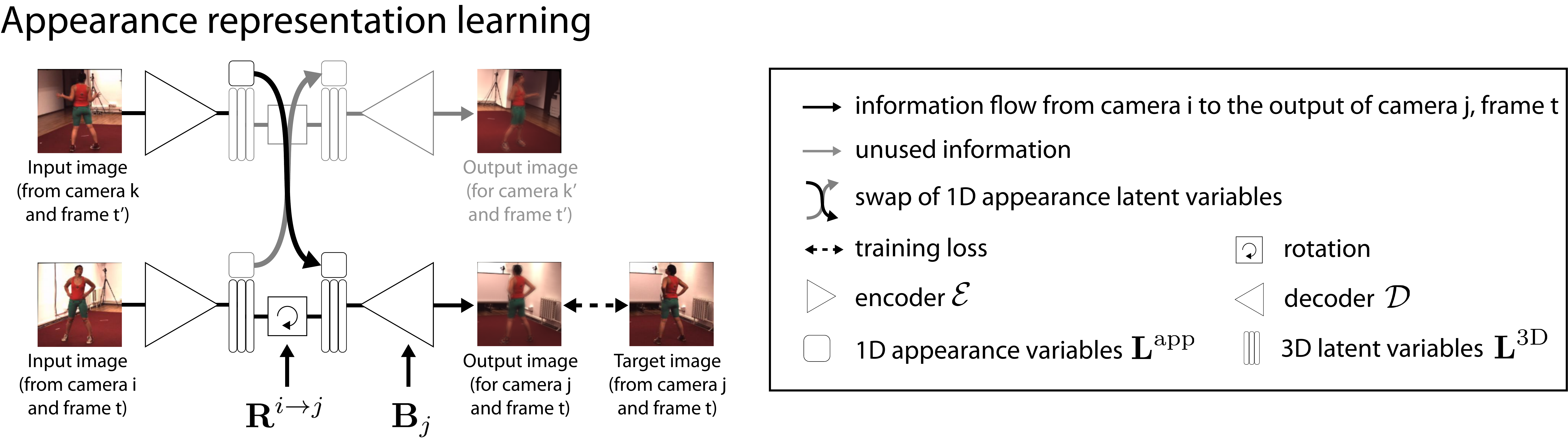}%
\vspace{-2mm}
\caption{{\bf Appearance representation learning.} To encode subject identity, we split the latent space into a 3D geometry part and an appearance part. The latter is not rotated, but swapped between two time frames $t$ and $t'$ depicting the same subject, so as to enforce it not to contain geometric information.
}
\label{fig:identityLearning}
\end{figure}

\paragraph{\bf Factoring out appearance.}

To separate appearance from geometry in our latent representation, we break up the output of the encoder $\cE$ into two separate vectors $\LatentG$ and $\LatentA$ that should describe pose and appearance, respectively. To enforce this separation, we train simultaneously on two frames $\mI_t$ and $\mI_{t'}$ depicting the same subject at different times, $t$ and $t'$, as depicted in Fig.~\ref{fig:identityLearning}. While the decoder uses $\LatentG_t$ and $\LatentG_{t'}$, as before, it swaps $\LatentA_t$ and $\LatentA_{t'}$. In other words, the decoder uses $\LatentG_t$ and $\LatentA_{t'}$ to resynthesize frame $t$ and $\LatentG_{t'}$ and $\LatentA_t$ for frame $t'$. Assuming that the person's appearance does not change drastically between $t$ and $t'$ and that differences in the images are caused by 3D pose changes, this results in  
$\LatentG$ encoding pose while $\LatentA$ encodes appearance. 

In practice, the encoder $\cE$ has two outputs, that is, $\cE_{\theta_e}: \mI_t^i \to (\LatentG_{i,t},\LatentA_{i,t})$ and the decoder $\cD_{\theta_d}$ accepts these plus the background as inputs, after swapping appearance and rotating the geometric representation for two views $i$ and $j$.
We therefore write the output of our encoder-decoder as
\begin{equation}
\cA_{\theta_e,\theta_d}(\mI_t^i, \mR^{i \to j},\ \LatentA_{k,t'},\ \LatentBG_{j}) =  \cD_{\theta_d}(\mR^{i \to j} \LatentG_{i,t},\ \LatentA_{k,t'},\ \LatentBG_{j}) \;.
\label{eq:combinedA}
\end{equation}

\paragraph{\bf Combined optimization.}

To train $\cA$ with sequences featuring several people and backgrounds, we randomly select mini-batches of $Z$ triplets $(\mI_t^i,\mI_t^j,\mI_{t'}^k)$ in $\cU$, with $t \neq t'$, from individual sequences. In other words, all three views feature the same person. The first two are taken at the same time but from different viewpoints. The third is taken at a different time and from an arbitrary viewpoint~$k$. For each such mini-batch, we compute the loss
\begin{equation}
E_{\theta_d,\theta_e} = \frac{1}{Z} \sum_{\substack{\mI_t^i,\mI_t^j,\mI_{t'}^k \in \cU\\t \neq t'}} \| \cA_{\theta_e,\theta_d}(\mI_t^i, \mR^{i \to j}, \LatentA_{k,t'},\ \LatentBG_{j}) - \mI_t^j \|^2 \;,
\label{eq:E}
\end{equation}
where %
$\Latent_{k,t'} = (\LatentG_{k,t'},\LatentA_{k,t'})$ is the output of encoder $\cE_{\theta_e}$ applied to image $\mI_{t'}^k$,
 $\LatentBG_{j}$ is the background in view $j$, and $\mR^{i \to j}$ denotes the rotation from view $i$ to view $j$. Note that we apply $\cE$ twice, to obtain $\LatentG_{i,t}$ and $\LatentA_{k,t'}$ in Eq.~\ref{eq:E} while ignoring 
$\LatentA_{i,t}$ and $\LatentG_{k,t'}$ with the swap discussed above. 

At training time, we minimize a total loss that is the sum of the pixel-wise error $E_{\theta_d,\theta_e}$ of Eq.~\ref{eq:E} and a second term obtained by first applying a Resnet with 18 layers trained on ImageNet on the output and target image and then computing the squared feature difference after the second block level, as previously done with VGG by~\cite{Park17}. All individual pixel and feature differences are averaged and their influence is balanced by weighting the feature loss by two. This additional term allows for crisper decodings and improved pose reconstruction.

\paragraph{\bf Rotation augmentation and cropping.} 

It is common practice for human pose estimation algorithms to first crop the subject of interest to factor out scale and global position, which are inherently ambiguous for monocular reconstruction and NVS. We also do this and use the crop information provided in the training datasets. More precisely,  we compute the rotation between two views with respect to the crop center instead of the image center and shear the cropped image so that it appears as if it were taken from a virtual camera pointing in the crop direction. We also apply random in-plane rotations to increase the diversity of the training set. As a result, in practice,  the rotation $\mR^{i \to j}$ and background crop $\mB_j$ depend on time $t$, but we neglect this dependency in our notation for readability.

\comment{
Our goal is to free the network from the task of predicting the target background because we are truly interested in modeling foreground objects, that is, a human in our case.
Furthermore, prediction of the background is not only unnecessary but also impossible without prior information. For instance for opposing views, the target background is at most partially visible.
To discard the background, we assume that the target background  is known during training and introduce a direct connection from $\mB$ to the decoder, as shown in Fig.~\ref{fig:representationLearning}. 
\MS{It seems strange to discuss this in Section 4, as was mentioned before. I would rather describe it here, as attempted in the following statement.} \ms{%
In practice, an approximation to $\mB$ is sufficient and easy to obtain. We use a static camera setup and compute the background image $\mB$ automatically by taking the median value of each pixel, \ms{for each} training sequence and camera. In some sequences, ghosting artifacts remain, but this form of noise did not seem to have a negative effect on training. 
In contrast to previous approaches~\cite{Zhou17unsupervised,Cohen14,Worrall17}, no precise background subtraction and silhouette information is needed. The model learns to segment the foreground since the direct connection frees it from the task of inferring the target background from the source image. 
\MS{Note that, since our ultimate goal is 3D pose estimation from the latent representation, and not Novel View Synthesis,}
\HR{I wanted to say that thanks to the way we handle the background, no BG is needed during testing.}
\hr{Note that, since the separation of foreground (encoder) and succeeding fusion of foreground and background (decoder) is learned during training,}
neither the source nor the target background needs not be known at test time; any background image, such as a wallpaper or simply a constant color, can be used, as shown in Fig.~\ref{fig:rep_bg}.
}}

\comment{
Appearance information is more naturally represented as a label, e.g. ``he is wearing a red shirt", instead of in 3D space. For this reason, we make the encoder $\cE$ produce two separate outputs  and , and
 enforce them to capture identity information and pose, respectively.
 \ms{To achieve this,  While the network for each frame makes use of its own pose representation $\LatentG$, as discussed before, the appearance representations $\LatentA_t$ and $\LatentA_{t'}$ are exchanged by the two networks.}
This precludes the network from storing pose, i.e., frame dependent, information in $\LatentA$ as this information would be useless after the exchange. On the other hand, the limited capacity of $\LatentG$ favors outsourcing appearance and identity, i.e., time-invariant person specific, information to $\LatentA$. Furthermore, we feed $\LatentA$ as a location invariant feature to the decoder, to further support the encoding of overall shape and appearance, instead of spatially varying geometry that shall be encoded by $\LatentG$. \MS{I don't understand the previous sentence, particularly the notion of location invariant and spatially varying in this context.}
In practice, examples of the same person can be selected without additional annotation by drawing from the same video.
A similar exchange of information has been performed before in~\cite{Reed15} for analogy transformations that enforce a feature map $a$ to be to $b$ as $c$ is to $d$. It is also related to the separation of facial identity, illumination and pose proposed in~\cite{Kulkarni15,Yang15weakly}.

\paragraph{\bf Combined optimization.}

To train $\cE$ and $\cD$ with sequence featuring several people and backgrounds, we randomly select mini-batches of $Z$ triplets $(\mI_t^i,\mI_t^j,\mI_{t'}^k)$ in $\cU$, with $t \neq t'$, from individual sequence. In other words, all three views feature the same person. The first two are taken at the same time but from different viewpoints. The third is taken at a different time and from an arbitrary viewpoint. For each such mini-batch we minimize the loss
\begin{equation}
E_{\theta_d,\theta_e} = \frac{1}{Z} \sum_{\substack{\mI_t^i,\mI_t^j,\mI_{t'}^k \in \cU\\t \neq l}} \| \cD_{\theta_d}(\mR^{i \to j} \LatentG_{i,t},\ \LatentA_{k,t'},\ \LatentBG_{j}) - \mI_t^j \|^2 \;,
\label{eq:E}
\end{equation}
with $\Latent_{i,t} = (\LatentG_{i,t},\LatentA_{i,t})$ and   $\Latent_{k,t'} = (\LatentG_{k,t'},\LatentA_{k,t'})$ are the output of encoder $\cE_{\theta_e}$ applied to images $\mI_i^t$ and $\mI_k^{t'}$ respectively, $\LatentBG_{j}$ is the background in view $j$, and $\mR^{i \to j}$ denotes the rotation from view $i$ to view $j$. Note that that we use $\LatentG_{i,t}$ and $\LatentA_{k,t'}$ is Eq.~\ref{eq:E} while ignoring 
$\LatentA_{i,t}$ and $\LatentG_{k,t'}$ to perform the switch discussed above.  
}
\comment{
\paragraph{\bf Rotation augmentation and cropping.} It is common for human pose estimation methods to crop the subject of interest before reconstruction. This factors out scale and global position estimation, which is inherently ambiguous for monocular reconstruction and NVS. For that, we use the crop information provided in the datasets. Instead, the subject extend could also be localized through simple motion cues.
To be precise, we compute the rotation between two views with respect to the crop center instead of the image center and shear the cropped image so that it appears as if it was taken from a virtual camera pointing in the crop direction.
Furthermore, we apply random rotation augmentation, around the optical axis of the camera, during training.
All of these steps make the rotation $\mR^{i \to j}$ and background crop $\mB_j$ dependent on time $t$, but we neglect this dependency in our notation to aid readability.

four frames from the same sequence, therefore depicting the same subject, with each pair of consecutive frames taken at the same time but from different viewpoints. This allows the  switching of $\LatentA$ between the first and second pair and the rotation of $\LatentG$ between simultaneous views. 
In practice, we randomly sample these mini-batches.

\MS{I would tend to remove the rest of this section, i.e., the next three sentences.}
In practice, $\Latent_{i,t} = (\LatentG_{i,t},\LatentA_{i,t})$ is a large vector and its components are extracted by splitting and reshaping $\LatentG_{i,t}$ to 3D.
Implementation details are given in Sec.~\ref{sec:implementation}.
We evaluate in Sec.~\ref{sec:evaluation} that these extensions to \cite{Cohen14} and \cite{Worrall17} are crucial for precise encoding, and lead to higher decoding accuracies in the semi-supervised setup. 
}

\section{3D Human Pose Estimation}
\label{sec:implementation}

Recall that our ultimate goal is to infer the 3D pose of a person from a monocular image. Since $\LatentG$ can be rotated and used to generate novel views, we are already part way there. Being a $3 \times N$ matrix, it can be understood as a set of $N$ 3D points, but these do not have any semantic meaning. However, in most practical applications, one has to infer a pre-defined representation, such as a skeleton with $K$ major human body joints, encoded as a vector $\vp \in \R^{3 K}$.

To instantiate such a representation, we need a mapping $\cF : \LatentG \to \R^{3K}$, which can be thought as a different decoder that reconstructs 3D poses instead of images. To learn it, we rely on supervision. However, as we will see in the results section, the necessary amount of human annotations is much smaller than what would have been required to learn the mapping directly from the images, as in many other recent approaches to human pose estimation. 

Let $\cL = \{(\mI_t, \vp_t)\}_{t=1}^{N_s}$ be a small set of $N_s$ labeled examples made of image pairs  and corresponding ground-truth 3D poses. We model $\cF$ as a deep network with parameters $\theta_f$. We train it by minimizing the objective function
\begin{equation}
F_{\theta_f} = \frac{1}{N_s} \sum_{t=1}^{N_s} \| \cF_{\theta_f}(\LatentG_t) - \vp_t \|\;, \text{ with } (\LatentG_{t}, \cdot) = \cE_{\theta_e}(\mI_t) \;.
\label{eq:loss_f}
\end{equation}
Because our latent representation $\LatentG$ already encodes human 3D pose and shape, $\cF$ can be implemented as a simple fully-connected neural network. Together with the encoder-decoder introduced in Section~\ref{sec:latent}, which is trained in an unsupervised manner, they form the semi-supervised setup depicted by Fig.~\ref{fig:teaser}(b). In other words, our unsupervised representation does a lot of the hard-work in the difficult task of lifting the image to a 3D representation, which makes the final mapping comparatively easy.

\section{Evaluation}
\label{sec:evaluation}

In this section, we first evaluate our approach on the task of 3D human pose estimation, which is our main target application, and show that our representation enables us to use far less  annotated training data than state-of-the-art approaches to achieve better accuracy. We then evaluate the quality of our latent space itself and show that it does indeed encode geometry, appearance, and background separately.  

\vspace{-3mm}
\paragraph{\bf Dataset.} We use the well-known Human3.6M (H36M) \cite{Ionescu14b} dataset. It is recorded in a calibrated multi-view studio and ground-truth human poses are available for all frames. This makes it easy to compare different levels of supervision, unsupervised, semi-supervised, or fully supervised. As in previous approaches \cite{Zhou17a,Rhodin18a,Tome17,Popa17,Martinez17,Mehta17b}, we use the bounding boxes provided with the dataset to extract cropped images.

\subsection{Semi-Supervised Human Pose Estimation}

\begin{figure}[t]
	\centering
\subfloat[]{\includegraphics[width=0.5\linewidth]{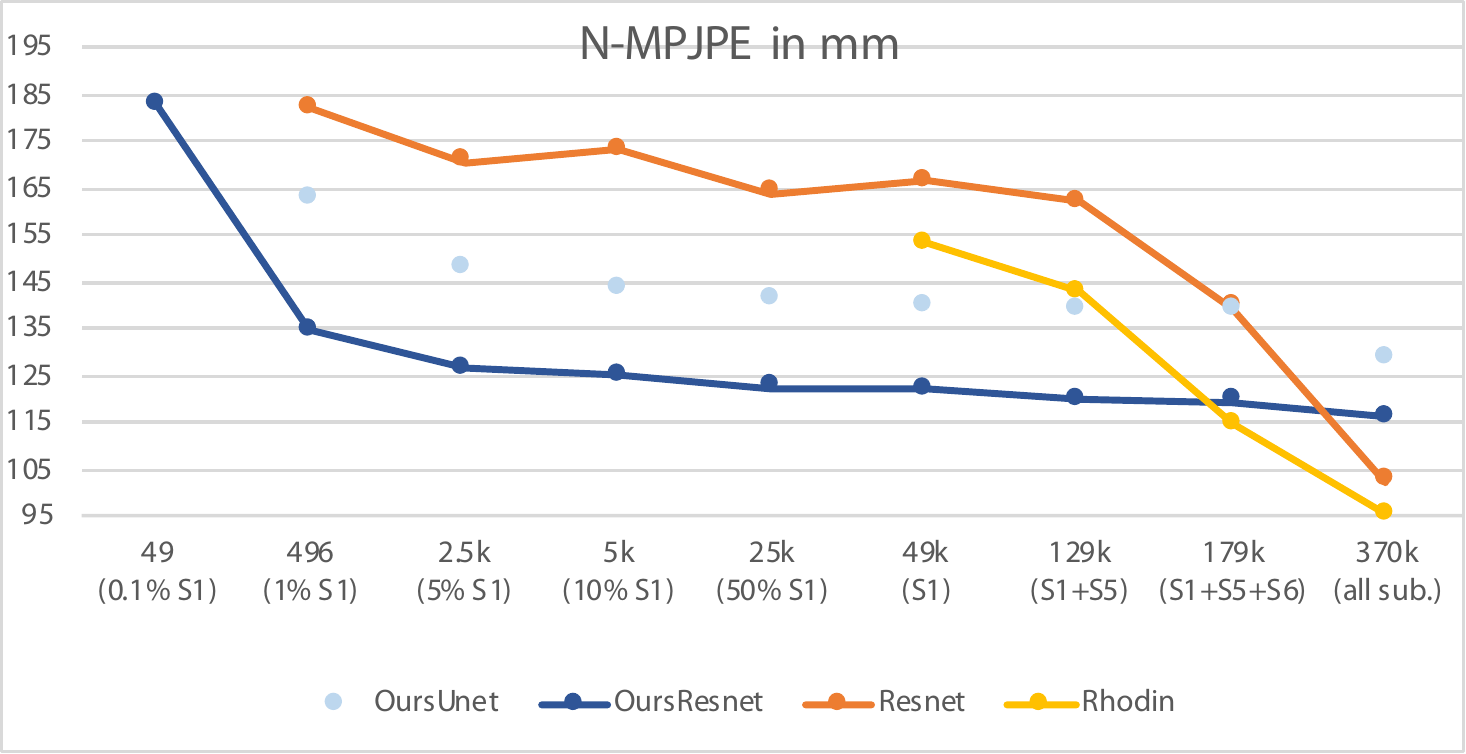}}%
\,%
\subfloat[]{\resizebox{0.45\columnwidth}{!}%
{%
	\begin{tabular}[b]{ |c|l|c|c|c|c| }%
		\hline
		Type & Method & MPJPE & NMPJPE& PMPJPE \\
		\hline
		Fully-supervised S1 & \resn{} & 177.2 & 166.5 & 135.6  \\
		\hline
		\multirow{3}{*}{Semi-supervised S1}
		& \rhod{}~\cite{Rhodin18a}\ & n/a & 153.3 & 128.6  \\
		& \unet{} & 149.5	& 135.9	& 106.4 \\
		& \rnet{} & {\bf 131.7} & {\bf 122.6} & {\bf 98.2}  \\
		\hline
	\end{tabular}%
}}\vspace{-2mm}
\caption{{\bf (a) Performance as function of the number of training samples.} When using all the available annotated 3D data in H36M, that is, 370,000 images, \rhod{} and \resn{} yield a better accuracy than our approach. However, when the number of training examples drops below 180'000 the baselines' accuracy degrades significantly, whereas \rnet{} degrades much more gracefully and our accuracy becomes significantly better. {\bf (b)} This improvement is consistent across the different metrics.}
	\label{fig:semiSupervised_quantitative}
\end{figure}

Our main focus is semi-supervised human pose estimation. We now demonstrate that, as shown in Fig.~\ref{fig:semiSupervised_quantitative}, recent state-of-the-art methods can do better than us when large amounts of annotated training data are available. However, as we use fewer and fewer of these annotations, the accuracy of the baselines suffers greatly whereas ours does not, which confers a significant advantage in situations where annotations are hard to obtain. We now explain in detail how the graphs of Fig.~\ref{fig:semiSupervised_quantitative} were produced and further discuss their meaning.

\vspace{-3mm}
\paragraph{\bf Metrics.}

We evaluate pose prediction accuracy in terms of the mean per joint prediction error (MPJPE), and its normalized variants N-MPJPE and P-MPJPE, where poses are aligned to the ground truth in the least-square sense either in scale only or in scale, rotation and translation, respectively,  before computing the MPJPE. The latter is also known as Procrustes alignment. We do this over 16 major human joints and all positions are centered at the pelvis, as in \cite{Zhou17a}. Our results are consistent across all metrics, as shown in Fig.~\ref{fig:semiSupervised_quantitative} (b).

\paragraph{\bf Baselines.} We compare our approach against the state-of-the-art semi-super\-vised method of~\cite{Rhodin18a}, which uses the same input as ours and outputs normalized poses. We will refer to it as \rhod{}. We also use the popular ResNet-based architecture~\cite{Mehta17b} to regress directly from the image to the 3D pose, as shown in  Fig.~\ref{fig:teaser}(c), we will refer to this as \resn{}.

Note that even higher accuracies on H36M than those of \rhod{} and \resn{} have been reported in the literature~\cite{Pavlakos16,Martinez17,Zhou17a,Tekin17a,Popa17} but they depend both on more complex architectures and using additional information such as labeled 2D poses~\cite{Mehta17b,Zhou17a,Tekin17a,Martinez17} or semantic segmentation~\cite{Popa17}, which is not our point here. 
What we want to show is that when using {\it only} 3D annotations and not many of them are available, our representation still allows us to perform well.

\vspace{-3mm}
\paragraph{\bf Implementation.}
\label{sec:implementation}

We base our encoder-decoder architecture on the UNet~\cite{Ronneberger15} network, which was used to perform a similar task in~\cite{Ma17}. We simply remove the skip connections to force the encoding of all information into the latent spaces and reduce the number of feature channels by half. 

Concretely the encoder $\cE$ consists of four blocks of two convolutions, where each two convolutions are followed by max pooling. The resulting convolutional features are of dimension {$512\times16\times16$} for an input image resolution of $128\times128$ pixels. These are mapped to $\LatentA \in \R^{128}$ and $\LatentG \in \R^{200\times3}$ by a single fully-connected layer followed by dropout with probability 0.3. The decoder $\cD$ maps $\LatentG$ to a feature map of dimension  {$(512-128)\times16\times16$} with a fully-connected layer followed by ReLU and dropout and duplicates $\LatentA$ to form a spatial uniform map of size {$128\times16\times16$}. These two maps are concatenated and then reconstructed by four blocks of two convolutions, where the first convolution is preceded by bilinear interpolation and all other pairs by up-convolutions. Each convolution is followed by batch-normalization and ReLU activation functions. We also experimented with a variant in which the encoder $\cE$ is an off-the shelf Resnet with fifty layers  \cite{He16}, pre-trained on ImageNet, and the decoder is the same as before. We will refer to these two versions as \unet{} and \rnet{}, respectively.

The pose decoder $\cF$ is a fully-connected network with two hidden layers of dimension 2048. The ground-truth poses in the least-squares loss of Eq.~\ref{eq:loss_f} are defined as root-centered 3D poses. Poses and images are normalized by their mean and standard deviation on the training set. We use mini-batches of size 16 and the Adam optimizer with learning rate ${10}^{-3}$
 for optimization of $\theta_e$, $\theta_d$~and~$\theta_f$. 

\vspace{-3mm}
\paragraph{\bf Dataset splits.}

On H36M, we take the unlabeled set $\cU$ used to learn our representation to be the complete training set---S1, S5, S6, S7 and S8, where SN refers to all sequences of the N$^{th}$ subject---but without the available 3D labels. To provide the required supervision to train the shallow network of Fig.~\ref{fig:teaser}(b), we then define several scenarios. 
\begin{itemize}%
\item  Fully supervised training with the 3D annotation of all five training subjects.
 \item We use all the 3D annotations for S1; S1 and S5; or S1, S5 and S6.
 \item We use only 50\%, 10\%, 5\%, 1\% or 0.1\% of the 3D annotations for S1.
 \end{itemize}%
 In all cases we used S9 and S11 for testing.  We subsampled the test and training videos at 10ps  to reduce redundancy and validation time. The resulting numbers of annotated images we used are shown along the x-axis of Fig.~\ref{fig:semiSupervised_quantitative}.

\paragraph{\bf Comparison to the state of the art.}

\rhod{} is the only method that is designed to leverage unlabeled multi-view footage without using a supplemental 2D dataset~\cite{Rhodin18a}.
\unet{} outperforms it significantly, e.g., on labeled subject S1 by 17.4mm (11.3\% relative improvement) and \rnet{} even attains a gain of 30.7mm (20 \% relative improvement).
The fact that the Resnet architecture, training procedure, and dataset split is the same for \rnet{} and \rhod{} evidences that this gain is due to our new way of exploiting the unlabeled examples, thus showing the effectiveness of learning a geometry-aware latent representation in an unsupervised manner.

\paragraph{\bf Discussion.}

As shown in Fig.~\ref{fig:semiSupervised_quantitative}, when more than 180,000 annotated images are used the baselines outperform us. However, their accuracy decreases rapidly when fewer are available and our approach then starts dominating. It only loses accuracy very slowly down to 5,000 images and still performs adequately given only 500.

\begin{table}[b!]
\center
\resizebox{0.560\columnwidth}{!}
{
	\begin{tabular}{ |l|c|c|c|c| }
		\hline
		Method & N-MPJPE& P-MPJPE\\
		\hline
		 \unet{}$^\star$ & {\bf 145.6} &	{\bf 112.2}\\
		\unet{}$^\star$, w/o appearance space, as in \cite{Cohen14,Worrall17} & 159.0 & 117.1\\	
		\unet{}$^\star$, w/o background handling, as in \cite{Cohen14,Worrall17}& 159.6 & 124.6\\	
		\unet{}$^\star$,  w/o 3D latent space, as in \cite{Tatarchenko15,Tatarchenko16} & 191.7 & 139.0  \\
		\hline
		\multicolumn{3}{l}{$^\star$ no rotation augmentation. Errors are reported in mm.}\\
	\end{tabular}
}
\resizebox{0.430\columnwidth}{!}
{
	\begin{tabular}{ |l|c|c|c|c| }
		\hline
		Method & N-MPJPE& P-MPJPE\\
		\hline
		\unet{}$^\star$ & {\bf 145.6} &	{\bf 112.2}\\
		\unet{}$^\star$, bilinear upsampling & 149.2 & 114.1  \\
		\unet{}$^\star$, w/o ImgNet loss & 154.1 & 118.7\\
		\unet{}$^\star$, $\cF$ with 1 hidden layer & 157.4 & 121.9 \\
		\hline
		\multicolumn{3}{l}{}\\
	\end{tabular}
}
\caption{{\bf Ablation study,} using S1 for semi-supervised training. The extensions to the NVS methods \cite{Tatarchenko15,Tatarchenko16} and \cite{Cohen14,Worrall17} as well as further model choices improve accuracy.}
\label{tb:ablation}
\end{table}

To better evaluate different aspects of our approach, we use the \unet{}  version to conduct an ablation study whose results we report in Table~\ref{tb:ablation}. In short, 
not separating the background and appearance latent spaces reduces N-MPJPE by 14 mm and P-MPJPE by more than 12 mm. Using two hidden layers in $\cF$ instead of one increases accuracy by 12 mm. The loss term based on ResNet-18 features not only leads to crisper NVS results but also improves pose estimation by 9 mm.  Using bilinear upsampling instead of deconvolution for all decoding layers reduces performance by 4 mm.  
The largest decrease in accuracy by far, 46.1mm, occurs when we use our standard \unet{} architecture but {\it without} our geometry-aware 3D latent space. It appears in the last line of the table on the left and strongly suggests that using our latent representation has more impact than tweaking the architecture in various ways. 

\subsection{Evaluating the Latent Representation Qualitatively}

We now turn to evaluating the quality of our latent representation as such with a number of experiments on \unet{}. We show that geometry can be separated from appearance and background and that this improves results. The quality of the synthesized images is best seen in the supplemental videos.

\vspace{-3mm}
\paragraph{\bf Novel View Synthesis.}

\begin{figure}[t]%
\captionsetup[subfigure]{labelformat=empty}
	\centering
	\subfloat[Input $i$]{\frame{\includegraphics[height=35px]{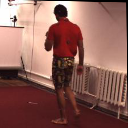}}}%
\,\,
\subfloat[View $j$]{\frame{\includegraphics[height=35px]{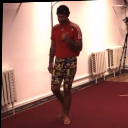}}}%
\subfloat[$\hdots$decoded]{\frame{\includegraphics[height=35px]{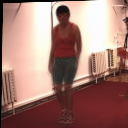}}}%
\,\,
\subfloat[View $j'$]{\frame{\includegraphics[height=35px]{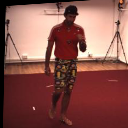}}}%
\subfloat[$\hdots$decoded]{\frame{\includegraphics[height=35px]{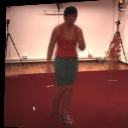}}}%
\,\,
\subfloat[View $j''$]{\frame{\includegraphics[height=35px]{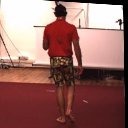}}}%
\subfloat[$\hdots$decoded]{\frame{\includegraphics[height=35px]{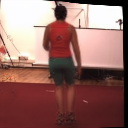}}}%
\\ \vspace{-0.2cm}
\centering
	\subfloat[0$^{\circ}$]{	 \includegraphics[height=35px]{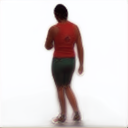}}%
\subfloat[45$^{\circ}$]{	\includegraphics[height=35px]{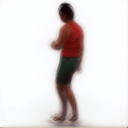}}%
\subfloat[90$^{\circ}$]{	\includegraphics[height=35px]{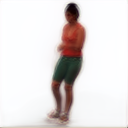}}%
\subfloat[135$^{\circ}$]{  \includegraphics[height=35px]{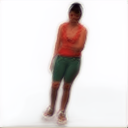}}%
\subfloat[180$^{\circ}$]{  \includegraphics[height=35px]{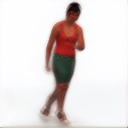}}%
\subfloat[225$^{\circ}$]{  \includegraphics[height=35px]{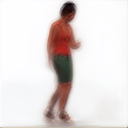}}%
\subfloat[270$^{\circ}$]{  \includegraphics[height=35px]{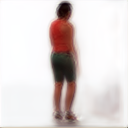}}%
\subfloat[315$^{\circ}$]{  \includegraphics[height=35px]{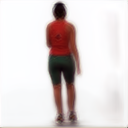}}%
        \vspace{-2mm}
	\caption{{\bf Novel viewpoint synthesis.} {\small {\bf Top row.} Input image from the training set to the left.  Each one of the three image pairs of image to the left of it comprise an original image acquired from a different viewpoint and the image synthesized from the input image. {\bf Bottom row.} We can also synthesize images for previously unseen viewpoints and remove the background. }}
\label{fig:rep_rotation}
\end{figure}

\begin{figure}[t!]
\captionsetup[subfigure]{labelformat=empty}
\centering
\subfloat[Input]{\frame{\includegraphics[height=45px]{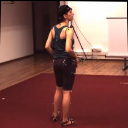}}}%
\,
\subfloat[Subject A]{%
	\includegraphics[height=45px]{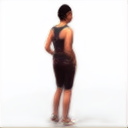}}%
\,
\subfloat[$\hdots$]{%
	\includegraphics[height=45px]{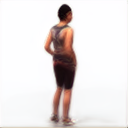}}%
\,
\subfloat[transition]{%
	\includegraphics[height=45px]{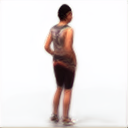}}%
\,
\subfloat[$\hdots$]{%
	\includegraphics[height=45px]{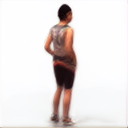}}%
\,
\subfloat[Subject B]{%
	\includegraphics[height=45px]{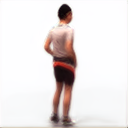}}%
\,
\subfloat[Target]{\frame{\includegraphics[height=45px]{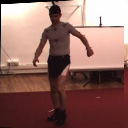}}}%
\\ \vspace{-0.4cm}
\subfloat[]{%
	\includegraphics[height=39px]{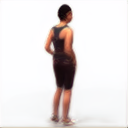}}%
\,
\subfloat[]{%
	\includegraphics[height=39px]{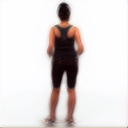}}%
\,
\subfloat[]{%
	\includegraphics[height=39px]{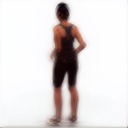}}%
\,
\subfloat[]{%
	\includegraphics[height=39px]{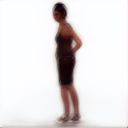}}%
\,
\subfloat[]{%
	\includegraphics[height=39px]{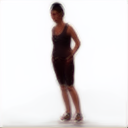}}%
\,
\subfloat[]{%
	\includegraphics[height=39px]{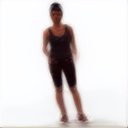}}%
\,
\subfloat[]{%
	\includegraphics[height=39px]{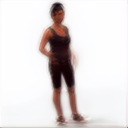}}%
\,
\subfloat[]{%
	\includegraphics[height=39px]{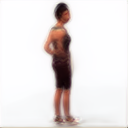}}%
\\ \vspace{-0.8cm}
\subfloat[0$^{\circ}$]{%
	\includegraphics[height=39px]{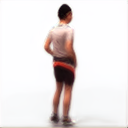}}%
\,
\subfloat[45$^{\circ}$]{%
	\includegraphics[height=39px]{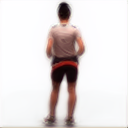}}%
\,
\subfloat[90$^{\circ}$]{%
	\includegraphics[height=39px]{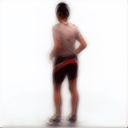}}%
\,
\subfloat[135$^{\circ}$]{%
	\includegraphics[height=39px]{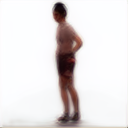}}%
\,
\subfloat[180$^{\circ}$]{%
	\includegraphics[height=39px]{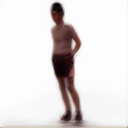}}%
\,
\subfloat[225$^{\circ}$]{%
	\includegraphics[height=39px]{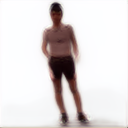}}%
\,
\subfloat[270]{%
	\includegraphics[height=39px]{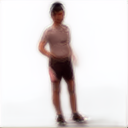}}%
\,
\subfloat[315$^{\circ}$]{%
	\includegraphics[height=39px]{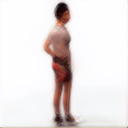}}%
\\ \vspace{-0.2cm}
\subfloat[Input]{\frame{\includegraphics[height=45px]{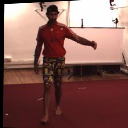}}}%
\,
\subfloat[Subject A]{%
	\includegraphics[height=45px]{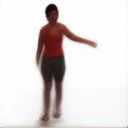}}%
\,
\subfloat[$\hdots$]{%
	\includegraphics[height=45px]{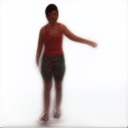}}%
\,
\subfloat[transition]{%
	\includegraphics[height=45px]{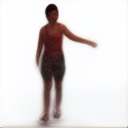}}%
\,
\subfloat[$\hdots$]{%
	\includegraphics[height=45px]{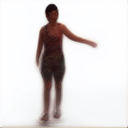}}%
\,
\subfloat[Subject B]{%
	\includegraphics[height=45px]{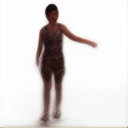}}%
\,
\subfloat[Target]{\frame{\includegraphics[height=45px]{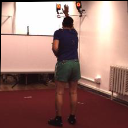}}}%
\\ \vspace{-0.4cm}
\subfloat[]{%
	\includegraphics[height=39px]{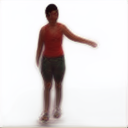}}%
\,
\subfloat[]{%
	\includegraphics[height=39px]{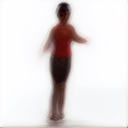}}%
\,
\subfloat[]{%
	\includegraphics[height=39px]{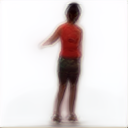}}%
\,
\subfloat[]{%
	\includegraphics[height=39px]{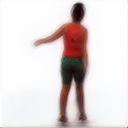}}%
\,
\subfloat[]{%
	\includegraphics[height=39px]{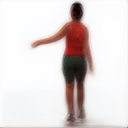}}%
\,
\subfloat[]{%
	\includegraphics[height=39px]{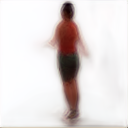}}%
\,
\subfloat[]{%
	\includegraphics[height=39px]{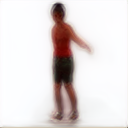}}%
\,
\subfloat[]{%
	\includegraphics[height=39px]{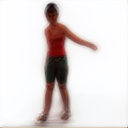}}%
\\ \vspace{-0.8cm}
\subfloat[0$^{\circ}$]{%
	\includegraphics[height=39px]{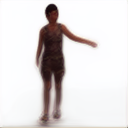}}%
\,
\subfloat[45$^{\circ}$]{%
	\includegraphics[height=39px]{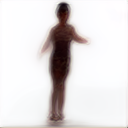}}%
\,
\subfloat[90$^{\circ}$]{%
	\includegraphics[height=39px]{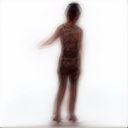}}%
\,
\subfloat[135$^{\circ}$]{%
	\includegraphics[height=39px]{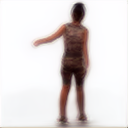}}%
\,
\subfloat[180$^{\circ}$]{%
	\includegraphics[height=39px]{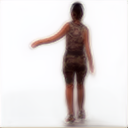}}%
\,
\subfloat[225$^{\circ}$]{%
	\includegraphics[height=39px]{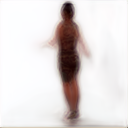}}%
\,
\subfloat[270$^{\circ}$]{%
	\includegraphics[height=39px]{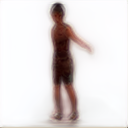}}%
\,
\subfloat[315$^{\circ}$]{%
	\includegraphics[height=39px]{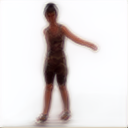}}%
\vspace{-2mm}
\caption{{\bf Appearance separation.} {\small The same pose can be decoded to different identities by blending appearance vectors in varying proportions. {\bf Top three rows.} We progressively replace the appearance of the input by that of the target and then generate rotated views, with both subjects appearing in the training set. {\bf Bottom three rows.} We do the same for subjects that appear in the test set. In all cases we set the background to white.}}
\label{fig:rep_appearance}
	\captionsetup[subfigure]{labelformat=empty}
	\centering
	\subfloat[]{\frame{\includegraphics[height=36px]{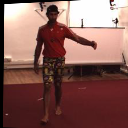}}}
\hfill
	\subfloat[]{\frame{\includegraphics[height=36px]{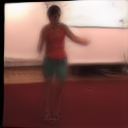}}}%
\hfill
	\subfloat[]{\frame{\includegraphics[height=36px]{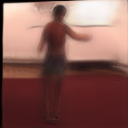}}}%
\hfill
	\subfloat[]{\frame{\includegraphics[height=36px]{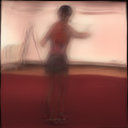}}}%
\hfill
	\subfloat[]{\frame{\includegraphics[height=36px]{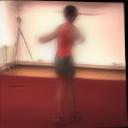}}}%
\hfill
	\subfloat[]{\frame{\includegraphics[height=36px]{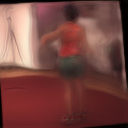}}}%
\hfill
	\subfloat[]{\frame{\includegraphics[height=36px]{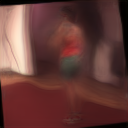}}}%
\hfill
	\subfloat[]{\frame{\includegraphics[height=36px]{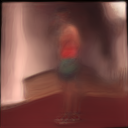}}}%
\hfill
	\subfloat[]{\frame{\includegraphics[height=36px]{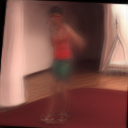}}}%
\\  \vspace{-0.8cm}
\centering
	\subfloat[Input ]{\frame{\includegraphics[height=36px]{images/NVS/testing_noBGskip/animation_i9_input.png}}}
\hfill
\subfloat[0$^\circ$]{{\includegraphics[height=36px]{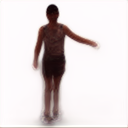}}}%
\hfill
\subfloat[45$^\circ$]{{\includegraphics[height=36px]{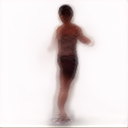}}}%
\hfill
\subfloat[90$^\circ$]{{\includegraphics[height=36px]{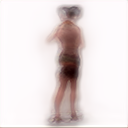}}}%
\hfill
\subfloat[135$^\circ$]{{\includegraphics[height=36px]{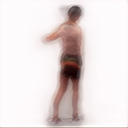}}}%
\hfill
\subfloat[180$^\circ$]{{\includegraphics[height=36px]{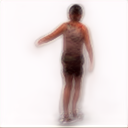}}}%
\hfill
\subfloat[225$^\circ$]{{\includegraphics[height=36px]{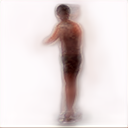}}}%
\hfill
\subfloat[270$^\circ$]{{\includegraphics[height=36px]{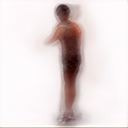}}}%
\hfill
\subfloat[315$^\circ$]{{\includegraphics[height=36px]{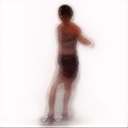}}}%
\vspace{-2mm}
	\caption{{\bf Ablation study.} {\small {\bf Top row.} Without background handling, as used in \cite{Cohen14,Worrall17}, the synthesized foreground pose appears fuzzy. {\bf Bottom row.} Without a geometry-aware latent space, as used by \cite{Tatarchenko15,Tatarchenko16},} results are inaccurate and blurred in new views. The input is the same as in Fig.~\ref{fig:rep_bg} and Fig.~\ref{fig:identityLearning}.}
	\label{fig:rep_nobg}
\end{figure}

Recall from Section~\ref{sec:latent} that $\cE$ encodes the image into variables $\LatentG$ and $\LatentA$, which are meant to represent geometry and appearance, respectively. To check that this is indeed the case, we multiply $\LatentG$ by different rotation matrices $\mR$ and feed the result along with the original $\LatentA$ to $\cD$, which produces novel views. Fig.~\ref{fig:rep_rotation} depicts such novel views when the input image belongs to the test.
Test subjects wear clothes that differ in color and shape from those seen in the training data. As a result, the geometry in the synthesized images remains correct, but the appearance ends up being a mixture of training appearances that approximates the unseen appearance. Arguably, using more than the five subjects that appear in the training set should result in a better encoding of appearance, which is something we plan to investigate in future work. 
Fig.~\ref{fig:rep_appearance} provides more examples, also on training subjects, where both the pose and appearance are accurately reproduced in the novel view, also for large rotations. 
For comparison purposes, in  Fig.~\ref{fig:rep_nobg}, we synthesize rotated images without using our geometry-aware latent space, that is, as in~\cite{Tatarchenko16}. The resulting images are far blurrier than those of Fig.~\ref{fig:rep_appearance}. Fig.~\ref{fig:rep_nobg} further shows that results degrade without the background handling, that is, as in \cite{Cohen14,Worrall17}.

\vspace{-1mm}
\paragraph{\bf Appearance and background switching.}

Let $\mI_j$ and $\mI_g$ be two images of subjects $j$ and $g$ and $(\LatentG_j,\LatentA_j,\LatentBG_j)=\cE(\mI_j)$ and $(\LatentG_g,\LatentA_g,\LatentBG_g)=\cE(\mI_g)$ their encodings. Re-encoding using $\LatentG$ of one and $\LatentA$ of the other yields results such as those depicted by  Fig.~\ref{fig:rep_appearance}. Note that the appearance of one is correctly transferred to the pose of the other while the geometry remains intact under rotation.
This method could be used to generate additional training data, by changing the appearance of an existing multi-view sequence to synthesize images of the same motion being performed by multiple actors.

\begin{figure}[t]%
\captionsetup[subfigure]{labelformat=empty}
	\centering
	\subfloat[]{\frame{\includegraphics[height=35px]{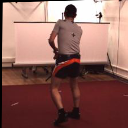}}}
	\,
\subfloat[]{\frame{\includegraphics[height=35px]{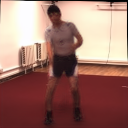}}}
	\,
\subfloat[]{\frame{\includegraphics[height=35px]{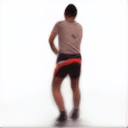}}}
	\,
\subfloat[]{\frame{\includegraphics[height=35px]{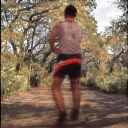}}}
	\,
\subfloat[]{\frame{\includegraphics[height=35px]{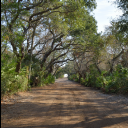}}}
\\ \vspace{-0.8cm}
\subfloat[Input]{\frame{\includegraphics[height=35px]{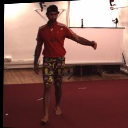}}}%
\,
\subfloat[New view]{\frame{\includegraphics[height=35px]{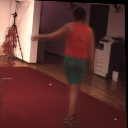}}}%
\,
\subfloat[White]{\frame{\includegraphics[height=35px]{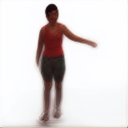}}}%
\,
\subfloat[Picture]{\frame{\includegraphics[height=35px]{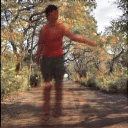}}}%
\,
\subfloat[Target]{\frame{\includegraphics[height=35px]{images/NVS/training_BG/selected/i13_random_bg.png}}}%
\vspace{-2mm}
\caption{{\bf Background separation.} {The background is handled separately from the foreground an can be chosen arbitrary at decoding time. From left to right, input image, decoded on the input background, on a novel view, on a white, and on a picture. The top row features someone from the training set and the bottom row from the test set.}}
\label{fig:rep_bg}
\end{figure}

Similarly, we can switch backgrounds instead of appearances before decoding the latent vectors, as shown in  Fig.~\ref{fig:rep_bg}. In one case, we make the background white and in the other we use a natural scene. In the first case,  dark patches are visible below the subject, evidently modeling shadowing effects that were learned implicitly. In the second case, the green trees tend to be rendered as orange because our training scenes were mostly reddish. Again, this is a problem a larger training database would almost certainly cure.

\begin{figure}[t]%
	\captionsetup[subfigure]{labelformat=empty}
	\centering
	\subfloat[Input ]{\frame{\includegraphics[height=36px]{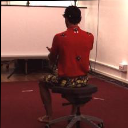}}}
\hfill
	\subfloat[0$^\circ$]{\frame{\includegraphics[height=36px]{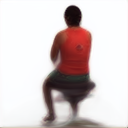}}}%
\hfill
	\subfloat[45$^\circ$]{\frame{\includegraphics[height=36px]{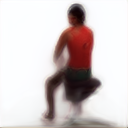}}}%
\hfill
	\subfloat[90$^\circ$]{\frame{\includegraphics[height=36px]{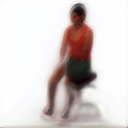}}}%
\hfill
	\subfloat[135$^\circ$]{\frame{\includegraphics[height=36px]{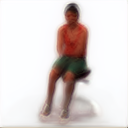}}}%
\hfill
	\subfloat[180$^\circ$]{\frame{\includegraphics[height=36px]{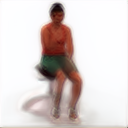}}}%
\hfill
	\subfloat[225$^\circ$]{\frame{\includegraphics[height=36px]{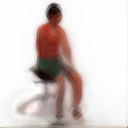}}}%
\hfill
	\subfloat[270$^\circ$]{\frame{\includegraphics[height=36px]{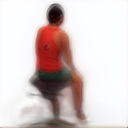}}}%
\hfill
	\subfloat[315$^\circ$]{\frame{\includegraphics[height=36px]{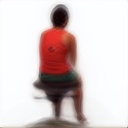}}}%
	\caption{{\bf Foreground objects} are reconstructed too, if seen in training and testing.}
	\label{fig:chair}
\end{figure}

\subsection{Generalization and Limitations}

In the data we used, some of the images contain a chair on which the subject sits. Interestingly, as shown in Fig.~\ref{fig:chair}, the chair appearance and 3D position is  faithfully reconstructed by our method. This suggests that it is not specific to human poses and can generalize to rigid objects. In future work, we therefore intend to apply it to more generic problems than human 3D pose estimation.

\begin{figure}[t]
	\captionsetup[subfigure]{labelformat=empty}
	\centering
\subfloat[]{\frame{
	\includegraphics[height=33px,trim=0 0 128 0,clip]{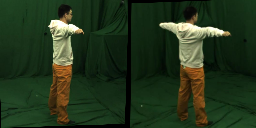}}}
\hfill
	\subfloat[]{\frame{\includegraphics[height=33px]{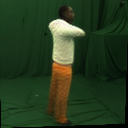}}}%
	\subfloat[]{\frame{\includegraphics[height=33px]{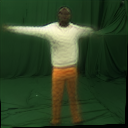}}}%
	\subfloat[]{\frame{\includegraphics[height=33px]{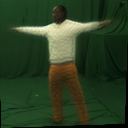}}}%
\hfill
	\subfloat[]{\frame{\includegraphics[height=33px]{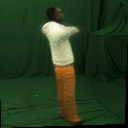}}}%
	\subfloat[]{\frame{\includegraphics[height=33px]{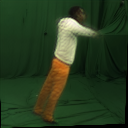}}}%
	\subfloat[]{\frame{\includegraphics[height=33px]{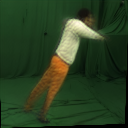}}}%
\hfill
	\subfloat[]{\frame{\includegraphics[height=33px]{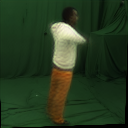}}}%
	\subfloat[]{\frame{\includegraphics[height=33px]{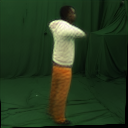}}}%
	\subfloat[]{\frame{\includegraphics[height=33px]{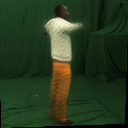}}}%
\\  \vspace{-0.8cm}
\centering
\subfloat[Input]{\frame{
	\includegraphics[height=33px,trim=0 0 128 0,clip]{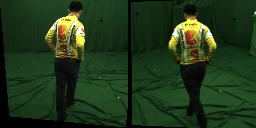}}}
\hfill
	\subfloat[]{\frame{\includegraphics[height=33px]{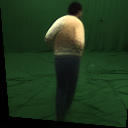}}}%
	\subfloat[yaw]{\frame{\includegraphics[height=33px]{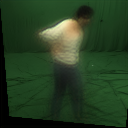}}}%
	\subfloat[]{\frame{\includegraphics[height=33px]{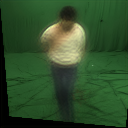}}}%
\hfill
	\subfloat[]{\frame{\includegraphics[height=33px]{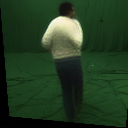}}}%
	\subfloat[roll]{\frame{\includegraphics[height=33px]{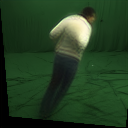}}}%
	\subfloat[]{\frame{\includegraphics[height=33px]{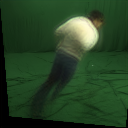}}}%
\hfill
	\subfloat[]{\frame{\includegraphics[height=33px]{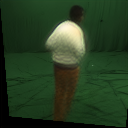}}}%
	\subfloat[pitch]{\frame{\includegraphics[height=33px]{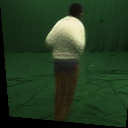}}}%
	\subfloat[]{\frame{\includegraphics[height=33px]{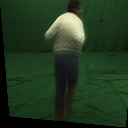}}}%
\vspace{-2mm}
	\caption{{\bf Generalization on 3DHP.} Our NVS solution generalizes well to the different camera placements in 3DHP, allowing for yaw, pitch and roll transformations.}
	\label{fig:3DHP_roll_pitch_yaw}
\end{figure}

We further tested our method on the MPI-INF-3DHP (3DHP)~\cite{Mehta17a} dataset, which features more diverse clothing and viewpoints, such as low-hanging and ceiling cameras, and is therefore well suited to probe extreme conditions for NVS. Without changing any parameter, \rnet{} is able to synthesis novel views in roll, yaw and pitch, as shown in Fig.~\ref{fig:3DHP_roll_pitch_yaw}. On H36M, only yaw and roll transformations work since the chest-height training views do not comprise any other viewpoint changes. The synthesized views on 3DHP are slightly more blurred than those of H36M. This is likely due to this dataset being significantly more difficult. It features different camera views, loose clothing, and more extreme motions and would call for a learning our latent representation using a larger multi-view training set.

\comment{
	\paragraph{Quantitative analysis---explicit 3D encoding.}
	\input{tex/tab_rotation}
	The NVS experiments described above suggest that our approach properly separates geometry to appearance. For a more qualitative assessment, we checked if, given two views $i$ et $j$, the known rotation from one to the other $\mR^{i \to j}$ is related to the rotation that best aligns the  corresponding vectors $\LatentG_i$ and $\LatentG_j$. In other words, we find the rotation $\hat{R}^{i \to j}$ that minimizes $\|\hat{R}^{i \to j} \LatentG_i - \LatentG_j\|$ and compute the distance between $\mR^{i \to j}$ and $\hat{R}^{i \to j}$ in \pf{terms of angular deviation between the two}. \PF{How do you measure this? Angular distance between the two rotations vectors?} We report the results in Tab.~\ref{tb:rotation} and compare them to those of for a standard auto-encoder. \PF{Which one?} The deviations are clearly much smaller in our case. \PF{But not all that small ...... Anyhow, wouldn't mutual information be a better measure?}
}

\comment{
	Because the difficulty task of lifting the image to a 3D geometric representation is learned entirely without 3D labels, our method is well suited for cases where much more unlabeled than labeled examples are available. We verify this by selecting different percentiles of the labels of S1. As shown in Fig.~\ref{fig:semiSupervised_quantitative}, if only 1\% of the labels are used, Our-ResNet still outperforms all baselines (\cite{Rhodin18a} with S1+S5, 142.9 mm vs. Our-ResNet 1\%-S1, 134.5 mm N-MPJPE). The same trend holds in terms of unnormalized MPJPE (\cite{Rhodin18a} with S1+S5, 173.6 mm vs. Our-ResNet 1\%-S1, 142.1 mm).
	Only when a lot of labels are available (S1+S5+S6), i.e., more than half of the training data is annotated, does the method of \cite{Rhodin18a} become superior. However, in the realm equally-sized labeled and unlabeled sets, the fully-supervised baseline is already very accurate and the absolute gain of any semi-supervised method is expected to be marginal.
}

\section{Conclusion}

We have introduced an approach to learning a geometry-aware representation of the human body in an unsupervised manner, given only multi-view imagery. Our experiments have shown that this representation is effective both as an intermediate one for 3D pose estimation and for novel view synthesis. For pose estimation, our semi-supervised approach performs much better than state-of-the-art methods when only very little annotated data is available. In future work, we will extend its range by learning an equivalent latent representation for much larger multi-view datasets but still in an unsupervised manner.

\comment{
\section{Limitations and Discussion}

Our method expects a tight crop around the image, which our implementation infers conveniently from the annotation available in the used datasets.
For static cameras, this step could be automated by simple background subtraction and by more advanced algorithms that take the camera geometry into account.
We did not test the method on moving cameras, where the automatic extraction of relative camera orientation is difficult.

Some of the training and test images contain a chair on which the subject sits. Interestingly, the chair appearance and 3D position is reconstructed by our method faithfully. The method is not specific to reconstructing human pose and generalizes to rigid objects.
In future work, we will apply the method on purely rigid scenes and a single moving camera.
In this setup, the camera position can be obtained with existing SLAM approaches or could be inferred alongside optimization \cite{}.

Extension to volumetric grids (beyond occupancy grids)

\begin{figure}[t]%
	\captionsetup[subfigure]{labelformat=empty}
	\centering
	\subfloat[Input ]{\frame{\includegraphics[height=36px]{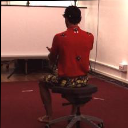}}}
\hfill
	\subfloat[0$^\circ$]{\frame{\includegraphics[height=36px]{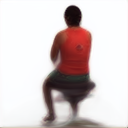}}}%
\hfill
	\subfloat[45$^\circ$]{\frame{\includegraphics[height=36px]{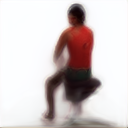}}}%
\hfill
	\subfloat[90$^\circ$]{\frame{\includegraphics[height=36px]{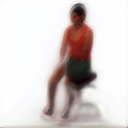}}}%
\hfill
	\subfloat[135$^\circ$]{\frame{\includegraphics[height=36px]{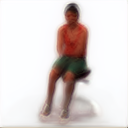}}}%
\hfill
	\subfloat[180$^\circ$]{\frame{\includegraphics[height=36px]{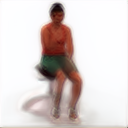}}}%
\hfill
	\subfloat[225$^\circ$]{\frame{\includegraphics[height=36px]{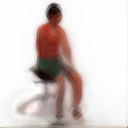}}}%
\hfill
	\subfloat[270$^\circ$]{\frame{\includegraphics[height=36px]{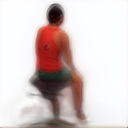}}}%
\hfill
	\subfloat[315$^\circ$]{\frame{\includegraphics[height=36px]{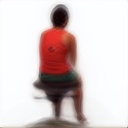}}}%
	\caption{{\bf Foreground objects} are reconstructed too, if seen in training and testing.}
	\label{fig:chair}
\end{figure}

The method is designed to handle the difficult articulated pose of human motion but it generalizes to rigid objects too. As evident in Fig.~\ref{fig:chair} where the present chair is reconstructed alongside the human.
}

\bibliographystyle{splncs}
\bibliography{../../../../bibtex/string,../../../../bibtex/vision,../../../../bibtex/learning,../../../../bibtex/biomed}

\begin{thebibliography}{10}

\bibitem{Mehta17a}
Mehta, D., Rhodin, H., Casas, D., Fua, P., Sotnychenko, O., Xu, W., Theobalt,
  C.:
\newblock {Monocular 3D Human Pose Estimation in the Wild Using Improved CNN
  Supervision}.
\newblock In: International Conference on 3D Vision. (2017)

\bibitem{Zhou17a}
Zhou, X., Huang, Q., Sun, X., Xue, X., We, Y.:
\newblock {Weakly-Supervised Transfer for 3D Human Pose Estimation in the
  Wild}.
\newblock arXiv Preprint (2017)

\bibitem{Pavlakos17}
Pavlakos, G., Zhou, X., Konstantinos, K.D.G., Kostas, D.:
\newblock {Harvesting Multiple Views for Marker-Less 3D Human Pose
  Annotations}.
\newblock In: Conference on Computer Vision and Pattern Recognition. (2017)

\bibitem{Tung17self}
Tung, H.Y., Tung, H.W., Yumer, E., Fragkiadaki, K.:
\newblock Self-supervised learning of motion capture.
\newblock In: Advances in Neural Information Processing Systems. (2017)
  5242--5252

\bibitem{Rhodin18a}
Rhodin, H., Spoerri, J., Katircioglu, I., Constantin, V., Meyer, F., Moeller,
  E., Salzmann, M., Fua, P.:
\newblock {Learning Monocular 3D Human Pose Estimation from Multi-View Images}.
\newblock In: Conference on Computer Vision and Pattern Recognition. (2018)

\bibitem{Zhou17unsupervised}
Zhou, X., Karpur, A., Gan, C., Luo, L., Huang, Q.:
\newblock Unsupervised domain adaptation for 3d keypoint prediction from a
  single depth scan.
\newblock arXiv preprint arXiv:1712.05765 (2017)

\bibitem{Yan16}
Yan, X., Yang, J., Yumer, E., Guo, Y., Lee, H.:
\newblock {Perspective Transformer Nets: Learning Single-View 3D Object
  Reconstruction Without 3D Supervision}.
\newblock In: Advances in Neural Information Processing Systems.
\newblock (2016)  1696--1704

\bibitem{Tulsiani17}
Tulsiani, S., Zhou, T., Efros, A., Malik, J.:
\newblock Multi-view supervision for single-view reconstruction via
  differentiable ray consistency.
\newblock In: Conference on Computer Vision and Pattern Recognition. Volume~1.
  (2017) ~3

\bibitem{Tatarchenko15}
Tatarchenko, M., Dosovitskiy, A., Brox, T.:
\newblock Single-view to multi-view: Reconstructing unseen views with a
  convolutional network.
\newblock CoRR abs/1511.06702 \textbf{1} (2015) ~2

\bibitem{Tatarchenko16}
Tatarchenko, M., Dosovitskiy, A., Brox, T.:
\newblock Multi-view 3d models from single images with a convolutional network.
\newblock In: European Conference on Computer Vision, Springer (2016)  322--337

\bibitem{Park17}
Park, E., Yang, J., Yumer, E., Ceylan, D., Berg, A.:
\newblock {Transformation-Grounded Image Generation Network for Novel 3D View
  Synthesis}.
\newblock In: Conference on Computer Vision and Pattern Recognition. (2017)
  702--711

\bibitem{Ionescu14a}
Ionescu, C., Papava, I., Olaru, V., Sminchisescu, C.:
\newblock {{Human3.6M}: Large Scale Datasets and Predictive Methods for 3D
  Human Sensing in Natural Environments}.
\newblock IEEE Transactions on Pattern Analysis and Machine Intelligence (2014)

\bibitem{Pavlakos16}
Pavlakos, G., Zhou, X., Derpanis, K., Konstantinos, G., Daniilidis, K.:
\newblock {Coarse-To-Fine Volumetric Prediction for Single-Image 3D Human
  Pose}.
\newblock In: Conference on Computer Vision and Pattern Recognition. (2017)

\bibitem{Tome17}
Tome, D., Russell, C., Agapito, L.:
\newblock {Lifting from the Deep: Convolutional 3D Pose Estimation from a
  Single Image}.
\newblock In: arXiv preprint, arXiv:1701.00295. (2017)

\bibitem{Popa17}
Popa, A.I., Zanfir, M., Sminchisescu, C.:
\newblock {Deep Multitask Architecture for Integrated 2D and 3D Human Sensing}.
\newblock In: Conference on Computer Vision and Pattern Recognition. (2017)

\bibitem{Martinez17}
Martinez, J., Hossain, R., Romero, J., Little, J.:
\newblock {A Simple Yet Effective Baseline for 3D Human Pose Estimation}.
\newblock In: International Conference on Computer Vision. (2017)

\bibitem{Mehta17b}
Mehta, D., Sridhar, S., Sotnychenko, O., Rhodin, H., Shafiei, M., Seidel, H.,
  Xu, W., Casas, D., Theobalt, C.:
\newblock {Vnect: Real-Time 3D Human Pose Estimation with a Single RGB Camera}.
\newblock In: ACM SIGGRAPH. (2017)

\bibitem{Rogez17}
Rogez, G., Weinzaepfel, P., Schmid, C.:
\newblock {Lcr-Net: Localization-Classification-Regression for Human Pose}.
\newblock In: Conference on Computer Vision and Pattern Recognition. (2017)

\bibitem{Tekin17a}
Tekin, B., M{\'{a}}rquez{-}neila, P., Salzmann, M., Fua, P.:
\newblock {Learning to Fuse 2D and 3D Image Cues for Monocular Body Pose
  Estimation}.
\newblock In: International Conference on Computer Vision. (2017)

\bibitem{Ionescu14b}
Ionescu, C., Carreira, J., Sminchisescu, C.:
\newblock {Iterated Second-Order Label Sensitive Pooling for 3D Human Pose
  Estimation}.
\newblock In: Conference on Computer Vision and Pattern Recognition. (2014)

\bibitem{Rhodin16}
Rhodin, H., Richardt, C., Casas, D., Insafutdinov, E., Shafiei, M., Seidel,
  H.P., B, S., Theobalt, C.:
\newblock {Egocap: Egocentric Marker-Less Motion Capture with Two Fisheye
  Cameras}.
\newblock ACM SIGGRAPH Asia \textbf{35}(6) (2016)

\bibitem{Rogez16}
Rogez, G., Schmid, C.:
\newblock {Mocap Guided Data Augmentation for 3D Pose Estimation in the Wild}.
\newblock In: Advances in Neural Information Processing Systems. (2016)

\bibitem{Chen16}
Chen, W., Wang, H., Li, Y., Su, H., Wang, Z., Tu, C., Lischinski, D., Cohen-or,
  D., Chen, B.:
\newblock {Synthesizing Training Images for Boosting Human 3D Pose Estimation}.
\newblock In: 3DV. (2016)

\bibitem{Varol17}
Varol, G., Romero, J., Martin, X., Mahmood, N., Black, M., Laptev, I., Schmid,
  C.:
\newblock Learning from synthetic humans.
\newblock In: Conference on Computer Vision and Pattern Recognition. (2017)

\bibitem{Joo15}
Joo, H., Liu, H., Tan, L., Gui, L., Nabbe, B., Matthews, I., Kanade, T.,
  Nobuhara, S., Sheikh, Y.:
\newblock {Panoptic Studio: A Massively Multiview System for Social Motion
  Capture}.
\newblock In: International Conference on Computer Vision. (2015)

\bibitem{Tulsiani18}
Tulsiani, S., Efros, A., Malik, J.:
\newblock {Multi-View Consistency as Supervisory Signal for Learning Shape and
  Pose Prediction}.
\newblock arXiv Preprint (2018)

\bibitem{Kar17}
Kar, A., H{\"a}ne, C., Malik, J.:
\newblock Learning a multi-view stereo machine.
\newblock In: Advances in Neural Information Processing Systems. (2017)
  364--375

\bibitem{Rezende16}
Rezende, D., Eslami, S., Mohamed, S., Battaglia, P., Jaderberg, M., Heess, N.:
\newblock {Unsupervised Learning of 3D Structure from Images}.
\newblock In: Advances in Neural Information Processing Systems. (2016)
  4996--5004

\bibitem{Zhu17}
Zhu, J.Y., Park, T., Isola, P., Efros, A.:
\newblock Unpaired image-to-image translation using cycle-consistent
  adversarial networks.
\newblock arXiv preprint arXiv:1703.10593 (2017)

\bibitem{Gadelha16}
Gadelha, M., Maji, S., Wang, R.:
\newblock 3d shape induction from 2d views of multiple objects.
\newblock arXiv preprint arXiv:1612.05872 (2016)

\bibitem{Grant16}
Grant, E., Kohli, P., van M.~Gerven:
\newblock {Deep Disentangled Representations for Volumetric Reconstruction}.
\newblock In: European Conference on Computer Vision. (2016)  266--279

\bibitem{Shu17a}
Shu, Z., Yumer, E., Hadap, S., Sunkavalli, K., Shechtman, E., Samaras, D.:
\newblock {Neural Face Editing with Intrinsic Image Disentangling}.
\newblock In: Conference on Computer Vision and Pattern Recognition. (2017)

\bibitem{Kim17}
Kim, H., Zollh{\"o}fer, M., Tewari, A., Thies, J., Richardt, C., Theobalt, C.:
\newblock {Inversefacenet: Deep Single-Shot Inverse Face Rendering from a
  Single Image}.
\newblock arXiv Preprint (2017)

\bibitem{Zhao17}
Zhao, B., Wu, X., Cheng, Z.Q., Liu, H., Feng, J.:
\newblock Multi-view image generation from a single-view.
\newblock arXiv preprint arXiv:1704.04886 (2017)

\bibitem{Bas17}
Bas, A., Huber, P., Smith, W., Awais, M., Kittler, J.:
\newblock {3D Morphable Models as Spatial Transformer Networks}.
\newblock arXiv Preprint (2017)

\bibitem{Tewari17}
Tewari, A., Zollh{\"o}fer, M., Kim, H., Garrido, P., Bernard, F., Perez, P.,
  Theobalt, C.:
\newblock Mofa: Model-based deep convolutional face autoencoder for
  unsupervised monocular reconstruction.
\newblock In: The IEEE International Conference on Computer Vision (ICCV).
  Volume~2. (2017)

\bibitem{Dosovitskiy15}
Dosovitskiy, A., Springenberg, J., Brox, T.:
\newblock {Learning to Generate Chairs with Convolutional Neural Networks}.
\newblock In: Conference on Computer Vision and Pattern Recognition. (2015)

\bibitem{Dosovitskiy17}
Dosovitskiy, A., Springenberg, J., Tatarchenko, M., Brox, T.:
\newblock {Learning to Generate Chairs, Tables and Cars with Convolutional
  Networks}.
\newblock IEEE Transactions on Pattern Analysis and Machine Intelligence
  \textbf{39}(4) (2017)  692--705

\bibitem{Chen16info}
Chen, X., Duan, Y., Houthooft, R., Schulman, J., Sutskever, I., Abbeel, P.:
\newblock {Infogan: Interpretable Representation Learning by Information
  Maximizing Generative Adversarial Nets}.
\newblock In: Advances in Neural Information Processing Systems. (2016)
  2172--2180

\bibitem{Tran17}
Tran, L., Yin, X., Liu, X.:
\newblock Disentangled representation learning gan for pose-invariant face
  recognition.
\newblock In: CVPR. Volume~3. (2017) ~7

\bibitem{Tung17}
Tung, H.Y., Harley, A., Seto, W., Fragkiadaki, K.:
\newblock Adversarial inverse graphics networks: Learning 2d-to-3d lifting and
  image-to-image translation from unpaired supervision.
\newblock In: The IEEE International Conference on Computer Vision (ICCV).
  Volume~2. (2017)

\bibitem{Kulkarni15}
Kulkarni, T.D., Whitney, W., Kohli, P., Tenenbaum, J.B.:
\newblock {Deep Convolutional Inverse Graphics Network}.
\newblock In: arXiv. (2015)

\bibitem{Thewlis17}
Thewlis, J., Bilen, H., Vedaldi, A.:
\newblock Unsupervised learning of object landmarks by factorized spatial
  embeddings.
\newblock In: International Conference on Computer Vision. (2017)

\bibitem{Thewlis17equivariant}
Thewlis, J., Bilen, H., Vedaldi, A.:
\newblock Unsupervised learning of object frames by dense equivariant image
  labelling.
\newblock In: Advances in Neural Information Processing Systems. (2017)
  844--855

\bibitem{Yang15weakly}
Yang, J., Reed, S., Yang, M.H., Lee, H.:
\newblock Weakly-supervised disentangling with recurrent transformations for 3d
  view synthesis.
\newblock In: Advances in Neural Information Processing Systems. (2015)
  1099--1107

\bibitem{Zhou16view}
Zhou, T., Tulsiani, S., Sun, W., Malik, J., Efros, A.:
\newblock View synthesis by appearance flow.
\newblock In: European conference on computer vision, Springer (2016)  286--301

\bibitem{Flynn16}
Flynn, J., Neulander, I., Philbin, J., Snavely, N.:
\newblock {Deepstereo: Learning to Predict New Views from the World's Imagery}.
\newblock In: Conference on Computer Vision and Pattern Recognition. (2016)
  5515--5524

\bibitem{Lassner17generative}
Lassner, C., Pons-Moll, G., Gehler, P.:
\newblock {A Generative Model of People in Clothing}.
\newblock arXiv Preprint (2017)

\bibitem{Ma17}
Ma, L., Jia, X., Sun, Q., Schiele, B., Tuytelaars, T., Gool, L.V.:
\newblock {Pose Guided Person Image Generation}.
\newblock In: Advances in Neural Information Processing Systems. (2017)
  405--415

\bibitem{Cohen14}
Cohen, T., Welling, M.:
\newblock {Transformation Properties of Learned Visual Representations}.
\newblock arXiv Preprint (2014)

\bibitem{Worrall17}
Worrall, D., Garbin, S., Turmukhambetov, D., Brostow, G.:
\newblock Interpretable transformations with encoder-decoder networks.
\newblock In: International Conference on Computer Vision. Volume~4. (2017)

\bibitem{Ronneberger15}
Ronneberger, O., Fischer, P., Brox, T.:
\newblock {{U-Net}: Convolutional Networks for Biomedical Image Segmentation}.
\newblock In: Conference on Medical Image Computing and Computer Assisted
  Intervention. (2015)

\bibitem{He16}
He, K., Zhang, X., Ren, S., Sun, J.:
\newblock {Deep Residual Learning for Image Recognition}.
\newblock In: Conference on Computer Vision and Pattern Recognition. (2016)
  770--778

\end{thebibliography}

\end{document}